\documentclass[conference]{IEEEtran}
\IEEEoverridecommandlockouts

\usepackage{cite}
\usepackage{hyperref}
\usepackage{amsmath,amssymb,amsfonts}
\usepackage{algorithmic}
\usepackage{graphicx}
\usepackage{textcomp}
\usepackage{xcolor}
\usepackage{amsmath, bm}
\usepackage{comment}

\ifCLASSOPTIONcompsoc
\usepackage[caption=false,font=normalsize,labelfon
t=sf,textfont=sf]{subfig}
\else
\usepackage[caption=false,font=footnotesize]{subfig}
\fi

\usepackage{soul,color,xcolor}
\newcommand{\jw}[1]{\sethlcolor{cyan}\hl{[Jennifer: #1]\\}}
\newcommand{\dave}[1]{\sethlcolor{orange}\hl{[David: #1]\\}}

\def\BibTeX{{\rm B\kern-.05em{\sc i\kern-.025em b}\kern-.08em
    T\kern-.1667em\lower.7ex\hbox{E}\kern-.125emX}}

\usepackage{fancyhdr}
\begin{document}

\title{Emotions as Ambiguity-aware\\Ordinal Representations{
}}

\author{
Jingyao Wu\textsuperscript{1*}\thanks{*Jingyao Wu is a fellow in the MIT-Novo Nordisk Artificial Intelligence Postdoctoral Fellows Program supported by funding from Novo Nordisk A/S.},
Matthew Barthet\textsuperscript{2},
David Melhart\textsuperscript{2},
Georgios N. Yannakakis\textsuperscript{2}\\
\emph{\textsuperscript{1}Media Lab, Massachusetts Institute of Technology, USA} \\
\emph{\textsuperscript{2}Institute of Digital Games, University of Malta, Msida, Malta} \\
jingyaow@mit.edu,\{matthew.barthet, david.melhart, georgios.yannakakis\}@um.edu.mt
}

\maketitle
\thispagestyle{fancy}
\begin{abstract}
Emotions are inherently ambiguous and dynamic phenomena, yet existing continuous emotion recognition approaches either ignore their ambiguity or treat ambiguity as an independent and static variable over time. Motivated by this gap in the literature, in this paper we introduce \emph{ambiguity-aware ordinal} emotion representations, a novel framework that captures both the ambiguity present in emotion annotation and the inherent temporal dynamics of emotional traces. Specifically, we propose approaches that model emotion ambiguity through its rate of change. We evaluate our framework on two affective corpora---RECOLA and GameVibe---testing our proposed approaches on both bounded (arousal, valence) and unbounded (engagement) continuous traces. 
Our results demonstrate that ordinal representations outperform conventional ambiguity-aware models on unbounded labels, achieving the highest Concordance Correlation Coefficient (CCC) and Signed Differential Agreement (SDA) scores, highlighting their effectiveness in modeling the traces' dynamics. For bounded traces, ordinal representations excel in SDA, revealing their superior ability to capture relative changes of annotated emotion traces. 
\end{abstract}


\section{Introduction}
Modeling emotions in a reliable fashion plays a critical role towards developing the next generation of human-centered artificial intelligence and human-machine interaction \cite{wang2022systematic}. Emotions in affective computing (AC) studies are traditionally represented either via discrete categories (e.g., happiness, sadness) \cite{ekman1999basic} or via continuous dimensions \cite{russell1980circumplex}. While discrete categories offer a straightforward interpretation, continuous representations, such as arousal, valence, or engagement allow for continuity both along the affect dimension considered and over time. 
Typically, continuous emotion labels (or traces) are collected from a group of human annotators tasked to perceive and annotate a series of stimuli, such as watching videos \cite{barthet2024gamevibe} or listening to speech samples \cite{ringeval2013introducing}.
Humans, however, perceive those stimuli differently due to a number of factors including gender, cultural background, personal experiences and reporting biases, leading to \emph{inter-rater ambiguity} \cite{sethu2019ambiguous}. For that purpose conventional Continuous Emotion Recognition (CER) systems \cite{russell1980circumplex,gunes2010automatic} often treat `ambiguity' as unwanted noise, and eliminate it by averaging multiple traces onto a derived \emph{gold standard} of the trace \cite{gunes2013categorical, ringeval2015avec, tzirakis2018end}. Ambiguity, however, reflects the subtle nature of emotions and highlights individual variations in perception, making it essential to account for when modeling emotion \cite{sethu2019ambiguous}. 

\begin{figure}[!tb]
\centering
\includegraphics[width=1.0\linewidth]{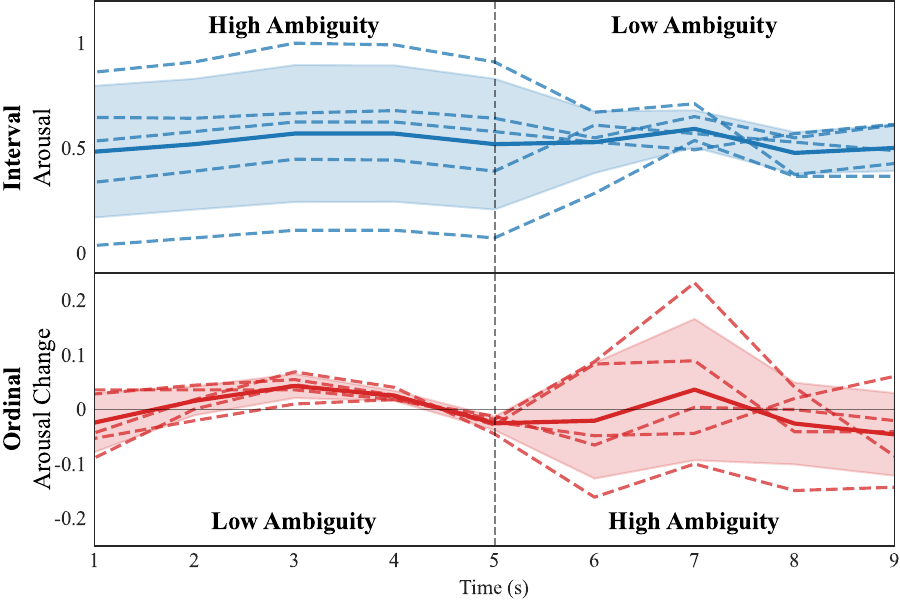}\vspace{-0.5em}
\caption{High-level conceptual visualization of ambiguity-aware \emph{interval}  vs. ambiguity-aware \emph{ordinal} representations for the same underlying data segment. Ambiguity is represented as an interval when the representation considers the absolute value of the arousal traces (top). Instead, it is represented as an ordinal phenomenon when one considers the change of the arousal traces over time (bottom). As seen from this contrastive example, the level of ambiguity is inverted (from high to low and inversely) when ambiguity is represented in an ordinal fashion. Dashed lines indicate multiple annotators, bold lines indicate the mean trace and shaded areas illustrate the spread of the trace distributions. 
} 
\label{fig:ratings}
\vspace{-3.3mm}
\end{figure}

Recently, there has been growing recognition of the importance of modeling ambiguity, leading to the development of CER models capable of predicting ambiguity-aware affect, which is represented as a probability distribution \cite{han2017hard, dang2018dynamic, bose2021parametric,wu2022novel, bose2024continuous, wu2024can, wu2024dual}. These approaches model traces at discrete time points and only treat them as interval labels, without considering any temporal aspects of ambiguity. Figure~\ref{fig:ratings} exemplifies the notion of \emph{ambiguity} by illustrating arousal traces solicited from multiple annotators (dashed lines) along with the associated ambiguity (shaded area). Under this formulation, the level of ambiguity is captured by the standard deviation of the trace distributions, assuming that a broad distribution at a single time step reflects high ambiguity, whereas a narrow distribution indicates low ambiguity.
However, the aforementioned assumption does not always hold, particularly when considering the trajectories of the traces. As observed in Fig. \ref{fig:ratings}, although the traces are widely spread (see top left figure) annotators follow a consistent trend and exhibit strong agreement on how affect (i.e., arousal in this example) evolves over time (see bottom left figure). Conversely, despite the low ambiguity across the traces (see top right figure), annotators showcase high levels of disagreement in the observed trace changes over time (see bottom right figure).

Such temporal variations in emotion perception can instead be captured through \emph{ordinal} representations \cite{yannakakis2018ordinal} (see bottom of Fig. \ref{fig:ratings}) that merely consider relative changes—--i.e., whether an emotional trace increases or decreases over time. Findings from a number of disciplines suggest that we tend to agree more on the temporal change of emotions rather than on their magnitude because we are naturally better at recognizing changes of such subjectively-defined constructs \cite{yannakakis2017ordinal, yannakakis2018ordinal, junge2013indirect,laming1984relativity, wu2023interval}. These observations suggest that current approaches to modeling ambiguity of emotions via \emph{interval} representations (see top of Fig. \ref{fig:ratings}) may not capture effectively the underlying information encoded in ambiguous emotion traces as they evolve over time. Consequently, in this paper, we address this challenge by modeling both the ambiguous and the ordinal nature of emotional constructs\footnote{Although strictly speaking \emph{ordinal} refers to non-measurable quantities with an underlying order (e.g., increase or decrease), in this paper we follow a second-order approach \cite{yannakakis2017ordinal,yannakakis2018ordinal} and use \emph{ordinal} to reflect the rate of change of the trace.}. For that purpose, we introduce \emph{Ambiguity-aware Ordinal Emotion Representations}, a novel approach that captures how emotions and their associated ambiguity evolve dynamically over time. Specifically, we propose and test two such distinct representation types:
\emph{group} representations that encode the temporal dynamics of affect by capturing how annotations evolve across multiple annotators 
and \emph{individual} representations that capture the temporal dynamics of each annotator's trace. 

By integrating the proposed emotion representations into conventional CER systems, we investigate the impact of ordinal modeling on ambiguity-aware emotion recognition. We hypothesize that (1) ordinal representations are particularly effective for modeling relative changes in emotion perception, and (2) incorporating ambiguity through probabilistic distributions better captures the temporal dynamics and inherent variability in human annotations. 
We present experiments on two well-established datasets, RECOLA \cite{ringeval2013introducing} and GameVibe \cite{barthet2024gamevibe}, testing the proposed approach on continuous affect traces that are given both as a bounded interval (RECOLA) and as unbounded ordinal changes (GameVibe) over time. Results obtained demonstrate that ambiguity-aware ordinal representations achieve superior performance in modeling continuous affective traces---particularly ordinal traces---highlighting their effectiveness in capturing relative changes in emotion perception. Furthermore, the \emph{group} representation consistently outperforms the \emph{individual} representation, emphasizing the benefits of representing annotators as a \emph{group} when modeling emotion in an ambiguity-aware ordinal fashion.

\section{Related Work}

\subsection{Modeling Emotion Ambiguity}\label{sec:background:ambiguity}

Ambiguity arises from the subjectivity of affective experiences \cite{yannakakis2018ordinal} and their diverse manifestations across individuals and contexts. In time-continuous affect labeling, ambiguity is often reflected in divergent annotator traces. Recent work has emphasized the need to model ambiguity explicitly, recognizing that disagreement among annotators is informative and should be integrated into affect recognition models to better capture the complexity of emotional understanding \cite{han2017hard, dang2018dynamic, bose2021parametric, wu2022novel, bose2024continuous, wu2024can, wu2024dual}. To achieve this, increasing efforts have been made to represent ambiguity using probability distributions. Various distribution types have proven effective, including parametric approaches (e.g., Gaussian distribution \cite{han2017hard}, Gaussian Mixture Model \cite{dang2018dynamic}, and Beta distribution \cite{bose2021parametric}) and non-parametric methods \cite{wu2022novel}. However, most existing studies view ambiguity as a function of emotion magnitude, overlooking the temporal dependencies in emotion and, thus, limiting the ability to fully capture its dynamic nature.

Among the few studies addressing temporal dependencies in emotion distributions, Dang et al., \cite{dang2018dynamic} modeled emotion labels using Gaussian mixture models and incorporated Kalman filters to capture the temporal evolution of the distributions' parameters, assuming that temporal dynamics follow a linear dynamical system. Several works have also adopted Long Short-Term Memory (LSTM)-based systems to learn temporal dependencies of distribution parameters \cite{han2017hard, bose2024continuous, wu2024can}. Wu et al. \cite{wu2022novel} explored the prediction of time-varying emotion distributions using a nonlinear dynamical system with a sequential Monte Carlo approach. More recently, a system with Dual--Constrained Dynamical Neural Ordinary Differential Equations ($\text{CD-NODE}_\gamma$) has been proposed to explicitly model changes of distribution parameters using ODE functions with additional constraints, ensuring smooth transitions and the validity of the distributions \cite{wu2024dual}. 

Despite the aforementioned  efforts to model temporal dynamics of emotion, existing research primarily assumes that ambiguity across raters is independent with respect to time, neglecting the ordinal nature of emotion annotations—--specifically, the consistency among raters in how emotions are annotated over time.

\subsection{Ordinal Affect Modeling}
\label{sec:background:ordinal}
Most current practices in soliciting time-continuous labels in affective computing predominantly rely on absolute ratings, while temporal dynamics and relative changes in ratings—key aspects of the relative ordinal nature of emotions—remain underexplored \cite{phillips2018impact}. Ordinal affect modeling explicitly acknowledges the relative nature of subjective emotional judgments, addressing the critical limitations of absolute rating-based approaches. 
Notably, treating ratings as absolute values often amplifies the inherent perception ambiguity of annotation signals. This is because people struggle to consistently map their internal emotional experiences onto a fixed scale. This, in turn, does not match how our internal value assessment system works, leading to personal perceptual ambiguity. Human affective assessments are inherently relative and subject to contextual factors and different anchoring and recency effects \cite{damasio1994descartes,seymour2008anchors,yannakakis2018ordinal}. Therefore, ordinal models offer a more robust representation, aligning better with our own cognitive processes when we annotate affect. Representing subjective labels in an ordinal fashion yields increased label consistency and inter-rater reliability across different disciplines including market research \cite{johnson2005relation}, behavioral economics \cite{tversky1981framing} and AC \cite{melhart2020study}.

\section{Ambiguity-aware Emotion Representation} \label{sec:method}

As discussed earlier, the existing dominant approach treats traces (that are inherently time-continuous) as interval data without accounting for any temporal change. We name this approach \emph{Interval Representation} and present it in Section \ref{sec:interval}. In contrast, we detail a new paradigm for modeling ambiguity-aware affect, capturing both the inherent ambiguity and the ordinal nature of annotated traces.  Our proposed types of \emph{Ordinal Representation} are detailed in Section \ref{sec:ordinal}. Since all representations considered are \emph{ambiguity-aware}, for simplicity purposes, we omit the term in the remainder of the paper. For a conceptual visualization of the two representations refer to Fig. \ref{fig:ratings}.

\subsection{Interval Representation} \label{sec:interval}

An increasing effort has been made in modeling ambiguity using probability distribution representations \cite{zhang2018dynamic, dang2018dynamic, bose2021parametric,wu2022novel, bose2024continuous, wu2024can, wu2024dual}. Under this formulation, given a set of ground truth annotations 
$\boldsymbol{y}_n = \{y_{n}^1, y_{n}^2, \dots, y_{n}^m\}$ from $m$ annotators at a single time instance $t_n$, for $n \in [1, N]$, the ambiguity-aware emotional state is assumed to be represented as the following probability distribution:

 \vspace{-0.5em}
\begin{equation}
     f(y; \boldsymbol{\theta}_n) = P(y|\boldsymbol{y}_n), \quad \boldsymbol{\theta} _n= \{\mu_n, \sigma_n\}
\end{equation}
where $f(y; \boldsymbol{\theta}_n)$ denotes a probabilistic model of the emotion state parameterized by $\boldsymbol{\theta}_n$;  $P(y|\boldsymbol{y}_n)$ represents the empirical distribution of ratings given the set of ground truth annotations $\boldsymbol{y}_n$ from multiple raters at time $t_n$; $\boldsymbol{\theta}_n = \{\mu_n, \sigma_n\}$ denotes the distribution parameters, where the central tendency $\mu_n$ of the distribution corresponds to the dominant emotional state and the standard deviation $\sigma_n$ reflects its associated ambiguity at $t_n$. 

In scenarios where emotion annotations are time-continuous traces, annotations from neighboring frames are often concatenated to incorporate the temporal information to enhance the fitting of the distribution. The concatenated annotations, including those from the neighboring frames, are denoted as:

\begin{equation}\label{eq:yn}
    \boldsymbol{\tilde{y}_n} = \{\boldsymbol{y}_{n-F}:\boldsymbol{y}_{n+F}\}
\end{equation}
where $F$ represents the number of neighboring frames considered around the current time instance $t_n$. Consequently, the final interval representation with the incorporation of neighboring information is given by:
\begin{equation}\label{eq:dist}
      f(y; \boldsymbol{\tilde{\theta}}_n) = P(y|\boldsymbol{\tilde{y}}_n), \quad \boldsymbol{\tilde{\theta}}_n = \{\tilde{\mu}_n, \tilde{\sigma}_n\}
\end{equation}
where $f(y; \boldsymbol{\tilde{\theta}}_n)$ denotes the probability distribution fitted with the set of concatenated ratings $\boldsymbol{\tilde{y}}_n$; $P(y|\boldsymbol{\tilde{y}}_n)$ represents the empirical distribution of ratings given $\boldsymbol{\tilde{y}}_n$ at time $t_n$. 

Note that we adopt the approach of concatenating neighboring frames in all experiments reported in this paper. Thus, for the sake of simplicity we disregard the tilde from all math notations; e.g., we use $\boldsymbol{\theta}_n$ instead of $\boldsymbol{\tilde{\theta}}_n$ in the remainder of the paper. 
The distribution parameters can be obtained using maximum likelihood estimation. 

The distributions used to fit the data can take various forms (e.g., Gaussian \cite{han2017hard}, Beta \cite{bose2024continuous}, non-parametric distribution \cite{wu2022novel} etc.); however, they are commonly represented using central tendency ($\mu$) and standard deviation ($\sigma$). Thus, in this paper, we denote the \emph{Interval Representation} as $I = \{\boldsymbol{\theta}_n\}, \forall n \in [1,N]$, where $\boldsymbol{\theta} _n= \{\mu_n, \sigma_n\}$. Fig. \ref{fig:representation:traces} depicts an example segment of arousal traces collected from the RECOLA dataset, with the associated ambiguity shown in the shaded area.

\subsection{Ordinal Representation}\label{sec:ordinal}

The proposed ordinal representation accounts for both genuine disagreement at each moment and consistent trends in how emotions evolve across multiple annotators' perceptions. Specifically, we propose two types of ordinal representation: \emph{Individual Representation}---denoted as $O^\text{I}$---
and \emph{Group Representation}---denoted as $O^\text{G}$---both of which are detailed in the remainder of this section.

\subsubsection{Individual Representation}

While the $I$ representation captures aspects of ambiguity, it overlooks any temporal information about the trace signal. To address this, $O^\text{I}$ aims to capture how an individual's perception of emotion evolves over time, as well as the variability across different individual annotators. It does so by first computing the rate of change (i.e., the trace's gradient) of each individual's trace, thereby capturing the unique temporal dynamics of each annotator. Then, it models the distribution of these gradients, enabling a population-level characterization of how different individuals annotate emotions. 

To derive $O^\text{I}$, we first compute the gradient of each annotator's ($m$) trace with respect to time, $g_n^m$, as follows:

\begin{equation}\label{eq:grad_ind}
g_n^m = \frac{dy_n^m}{dt} \quad \forall \quad  n \in [1,N]
\end{equation}
where 
$y_n^m$ denotes the trace of annotator $m$ at time $t_n$ and $N$ is the total number of time steps. 
These gradients are computed across past and future samples to ensure temporal smoothness, as defined in Eq. \eqref{eq:grad1}.

\begin{equation} \label{eq:grad1}
\frac{dx_n}{dt} \approx \frac{x_{n+1} - x_{n-1}}{2} 
\end{equation}
where $x$ is a time variable (e.g., annotator's traces $y^m$).

Then, we estimate a probability distribution $f(g_n^m; \boldsymbol{\phi}_n)$ over these gradients according to Eq. \eqref{eq:dist2}.  

\begin{equation} \label{eq:dist2}
f(g_n^m; \boldsymbol{\phi}_n), \quad \boldsymbol{\phi}_n = \{\mu_n^\text{I}, \sigma_n^\text{I}\}
\end{equation}
where $\boldsymbol{\phi}_n$ denotes the distribution parameters;  $\mu^\text{I}_n$ and $\sigma^\text{I}_n$ represent the central tendency and spread of the gradient distribution at time $t_n$, respectively. Finally, the \emph{individual} representation is given as $O^\text{I} = \{\boldsymbol{\phi}_n\},\forall n \in [1,N]$. 

Figure \ref{fig:representation:gradients} shows the trace gradients of multiple annotators 
illustrating the $\mu^\text{I}$ and $\sigma^\text{I}$ components of $O^\text{I}$ as derived from these gradients. 
The figure demonstrates how $O^\text{I}$ captures agreements (or disagreements) on arousal changes across multiple annotators. For instance, at around 24 seconds, the original traces appear to be flattened (see Fig. \ref{fig:representation:traces}); evidently, gradients at that time window are close to zero for all annotators (see Fig. \ref{fig:representation:gradients}), thereby yielding a very narrow distribution and centered around zero for $O^\text{I}$ (i.e., both $\mu^\text{I}$ and $\sigma^\text{I}$ are close to zero). In contrast, when individual gradients deviate substantially (e.g., between 5 and 10 seconds) the distribution to be modeled is wider.

\subsubsection{Group Representation}

The alternative \emph{group} ordinal representation aims to capture how the annotation distribution of the entire group of annotators changes over time. Specifically, we compute the gradients of the central tendency ($\mu$) and the ambiguity ($\sigma$) over time. This approach allows us to capture changes in the examined emotional state and its associated ambiguity. These changes provide a smooth representation of the temporal dynamics of the trace, accounting for subtle fluctuations.
Consequently, the \emph{group} representation is given as $O^\text{G} = \{g_n^\text{G}\},\forall n \in [1,N]$, where $g_n^\text{G}$ are the temporal gradients of the distribution's parameters (e.g., mean and standard deviation) at time $t_n$:

\vspace{-0.5em}
\begin{equation} \label{eq:grad_coh}
        g_n^\text{G} = \{\frac{d\mu_n}{dt}, \frac{d\sigma_n}{dt} \}, \forall n \in [1,N] 
\end{equation}
where $\{\mu_n, \sigma_n\}$ are the distribution parameters estimated according to Eq. \eqref{eq:dist}; $\frac{d\mu_n}{dt}$ and $\frac{d\sigma_n}{dt}$ denote, respectively, the rate of change of the distribution parameters $\mu_n$ and $\sigma_n$ at time $t_n$; these gradients are computed via Eq. \eqref{eq:grad1}.

\begin{figure}[!t]
\centering
\subfloat[Interval representation ($I$). The dashed lines show the trace of each annotator, the solid line shows the central tendency ($\mu$), and the shaded area represents the ambiguity ($\sigma$) of the trace.]{\includegraphics[width=\linewidth]{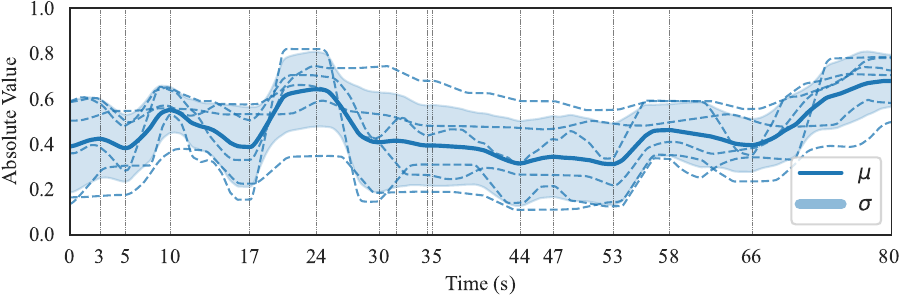}\label{fig:representation:traces}}

\hfil
\subfloat[Individual ordinal representation ($O^\text{I}$). The dashed lines depict the gradients of each annotator's trace as shown in Fig. \ref{fig:representation:traces}. The solid line and the shaded area represent, respectively, the central tendency of the gradients ($\mu^\text{I}$) and their standard distribution ($\sigma^\text{I}$).]{\includegraphics[width=\linewidth]{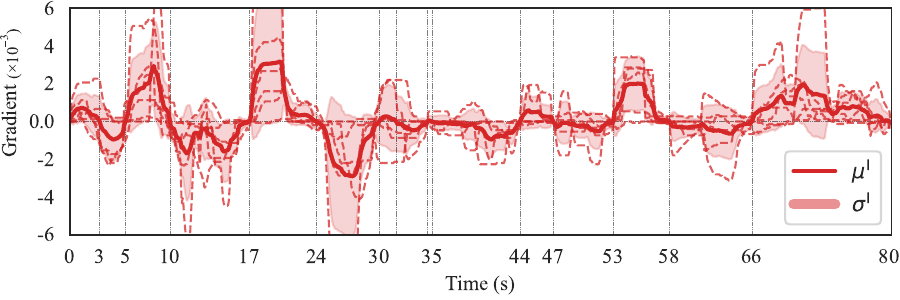}\label{fig:representation:gradients}}

\hfil
\subfloat[Group ordinal representation ($O^\text{G}$). The $d\mu/dt$ and $d\sigma/dt$ gradient traces are derived from $I$ ($\mu$ and $\sigma$, respectively) shown in Fig.~\ref{fig:representation:traces}.]{\includegraphics[width=\linewidth]{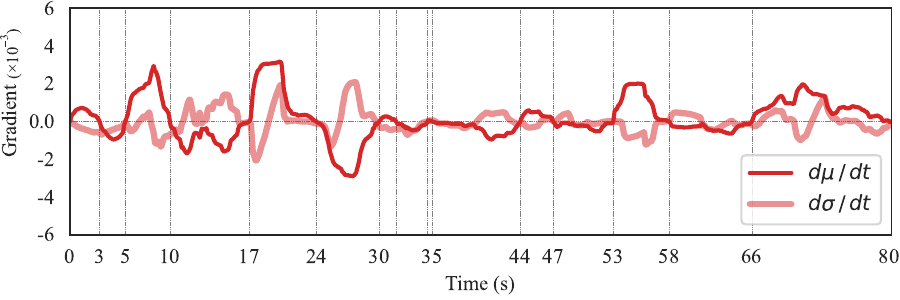}\label{fig:representation:group}}

\caption{An example of the different emotion representations as employed on RECOLA's arousal traces. The vertical lines divide the figures based on the tendency of the underlying data (Fig.~\ref{fig:representation:traces}) to increase or decrease. For the sake of visual simplicity in this example we set $F=0$ and show the traces without fitting them to a distribution.}\vspace{-2mm}
\label{fig:representation}
\end{figure}
Figure~\ref{fig:representation:group} illustrates the $\frac{d\mu}{dt}$ and $\frac{d\sigma}{dt}$ components of $O^\text{G}$. Comparing these traces to $I$ reveals how this representation amplifies relative changes. 
Observing the signals of Fig.~\ref{fig:representation:traces} between 17 and 24 seconds for instance 
we note a rise of $\mu$ accompanied by a lower level of $\sigma$---followed by a flattening of $\mu$ and a definite increase in $\sigma$.
In contrast in Fig.~\ref{fig:representation:group} ($O^\text{G}$), the $\frac{d\mu}{dt}$ trace rapidly rises and then falls close to zero as $\mu$ tapers off. Similarly $\frac{d\sigma}{dt}$ amplifies the temporal dynamics of the $\sigma$ trace observed in Fig.~\ref{fig:representation:traces}. Instead of a minor dip in values, with $\frac{d\sigma}{dt}$ we can observe a sharp decline, followed by a sharp increase encoding the rapidly changing levels of ambiguity before converging to values around zero (as the $\sigma$ trace settles around a constant value).

\section{Datasets}
Among the corpora that contain time-continuous emotion labels, the RECOLA and GameVibe datasets are selected for the experimental analysis in this paper. The RECOLA dataset has been widely used in continuous emotion recognition tasks and has been particularly employed in many recent studies on ambiguity modeling \cite{wu2022novel, bose2024continuous, dang2018dynamic, wu2024can, wu2024dual}. The GameVibe dataset is a newly introduced corpus of particular interest due to the ordinal nature of its emotion annotations.

\subsection{RECOLA Corpus}
The RECOLA dataset consists of 9.5 hours of spontaneous dyadic conversation recordings in French \cite{ringeval2013introducing}. The dataset is annotated by six human annotators with continuous arousal and valence traces within the range $[-1, 1]$. The original ratings are sampled at a period of 40 ms, which are further aggregated into a window size of 3 seconds. A 4 second time offset is applied to each utterance for both arousal and valence to compensate for the annotation delay \cite{huang2015investigation}. The training and development 
sets each contain nine five-minute utterances, consistent with the data partition in the AVEC challenge 2015 \cite{ringeval2015avec}. Since the challenge test sets are not publicly available, the system is trained, and hyperparameters are optimized using the training set and tested on the development set, as per standard practice \cite{wu2022novel, bose2024continuous, dang2018dynamic, wu2024dual}.

\subsection{GameVibe Corpus}

GameVibe \cite{barthet2024gamevibe} is a novel multimodal corpus consisting of 2 hours of gameplay video footage from 30 different first-person shooter games, annotated for viewer engagement by 20 annotators. The dataset is separated into 4 sessions containing 30 unique one minute clips---a clip from each game. Each session is annotated by the same set of 5 annotators; clips within each session are presented in a random order. The video clips are sampled at 30 Hz with a resolution of $1280\times720$ for modern games and $541\times650$ for older games. 

Annotations are collected using the RankTrace annotation tool \cite{lopes2017ranktrace} via the PAGAN \cite{melhart2019pagan} platform.
Annotators watched the game clips and provided engagement labels in real time by scrolling up or down on a mouse wheel to indicate increases or decreases in their engagement level respectively. No recordings of the participants' faces or the annotation interface form part of the final corpus. 
The engagement ratings are unbounded and originally sampled at a period of 250 ms and further aggregated into 3-second time windows following previous studies \cite{pinitas2024varying, pinitas2024across}. Some videos are slightly shorter than one minute. To maintain consistency, only the first 19 time windows are used, discarding any extras. One 10-window video is also excluded as an outlier due to its short length. Finally, the corpus is separated into train-validation splits using 10-fold cross-validation, resulting in a leave-3-games-out protocol. This means that the training sets for each fold consist of 108 videos from 27 games, and the validation sets consist 12 videos from 3 games.

\section{Experimental Settings} 
\label{sec:system}

\begin{figure*}[!tb]
\centering
\includegraphics[width=1.0\textwidth]{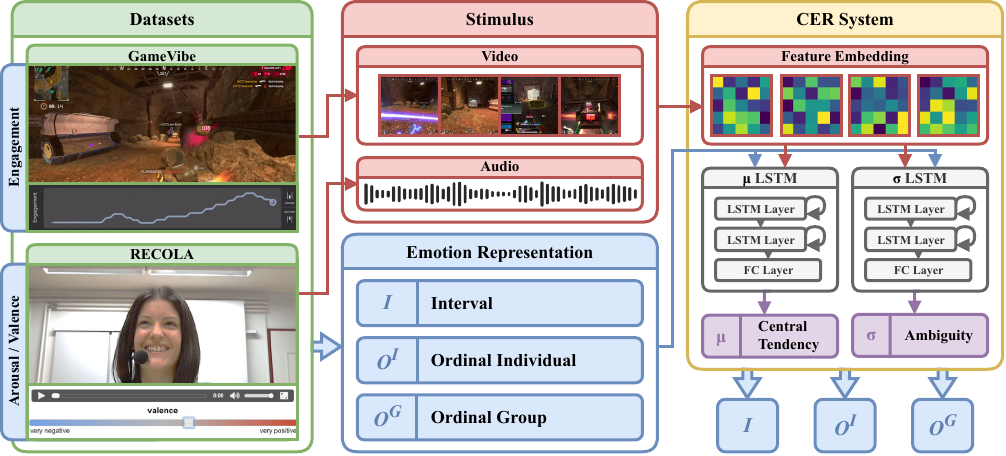}
\caption{A visual representation of the methodology followed. Traces from the RECOLA \cite{ringeval2013introducing} (arousal, valence) and GameVibe \cite{barthet2024gamevibe} (engagement) corpora are modeled using the three distinct \emph{emotion representations}. An LSTM-based \emph{continuous emotion recognition} system processes these representations along with two different emotional stimuli—--video from GameVibe and speech from RECOLA--—to generate time-continuous outputs over time for each emotion Representation.}
\label{fig:model}
\vspace{-1em}
\end{figure*}

In this section we integrate the proposed emotion representation into conventional continuous emotion recognition systems. We aim to investigate the impact of ordinal modeling on ambiguity-aware emotion recognition and assess whether representing emotions through their rate of change, rather than static values, enhances the system’s ability to align more closely with human perception---capturing both ambiguity and temporal dynamics.

A graphical representation of the framework is shown in Fig. \ref{fig:model}. The three different emotion representations detailed in Section \ref{sec:method} are tested as part of a CER system. This yields three types of model outputs as depicted in each box at the bottom of Fig. \ref{fig:model}: $I$, $O^{\text{G}}$ and $O^{\text{I}}$. Within each CER system, emotional stimuli (speech or video) are first fed into a feature extractor, which in turn feeds an LSTM-based architecture for CER. This follows the structure of the backbone model adopted in several recent studies demonstrating effectiveness in affect prediction \cite{zhang2018dynamic, bose2024continuous, wu2024can}. As each emotion representation consists of two separate parameters (i.e., $\mu$ and $\sigma$), we train two parameter-independent CER systems to optimize learning for both (see the two outputs illustrated in the magenta boxes of Fig. \ref{fig:model}. The feature extraction methods and distributions selected, the model implementation settings employed, and the evaluation measures used are detailed in the remainder of this section.

\subsection{Feature Extractor}

\subsubsection{Audio Features}

For emotion utterances from the RECOLA dataset, we adopt the Bag-of-Audio-Words (BoAW) features following state-of-the-art studies in ambiguity modeling in this dataset \cite{wu2024can, bose2024continuous}. The 20 extracted MFCCs and their deltas are first computed from the input speech utterances. Then, the audio words were determined as clusters in this space \cite{schmitt2016border}. The BoAW features employed in our experiments were generated using 100 clusters, yielding a 100-dimensional BoAW representation. Feature extraction was implemented using OpenXbow \cite{schmitt2017openxbow}, a detailed explanation of which can be found in \cite{schmitt2016border}. 

\subsubsection{Video Embeddings}
Visual features are extracted from GameVibe's videos using the VideoMAEv2 \cite{wang2023videomae}, a state-of-the-art pre-trained video masked autoencoder which has been used successfully on this corpus in earlier work \cite{pinitas2024varying}. Each video is processed as 3-second non-overlapping clips aligned with the time windows given for the annotation signals. We selected 16 frames from each of these clips by uniformly sampling across the three seconds to create a gif for each time window. We then used the base VideoMAEv2 model to extract embeddings for each gif, resulting in a latent vector of 768 values representing visual and temporal information across the time window. This resulted in a final training corpus of $2261\times768$ (time windows $\times$ embeddings).

\subsection{Distribution Selection}
We adopt Beta and Gaussian distributions to fit the distributions ($I$ representation) for the annotations in RECOLA and GameVibe datasets, respectively. For the arousal and valence traces of RECOLA---which are originally bounded within the $[-1, 1]$ range---we employ the Beta distribution given its suitability, compared to other commonly used distributions, in this dataset \cite{ringeval2013introducing}. The original annotations are first mapped to $[0,1]$ via a linear transformation in order to meet the requirement of Beta fit, following the approach described in \cite{bose2021parametric}. The annotations of neighboring frames $F = 1$ are concatenated for the estimation, according to Eq \eqref{eq:yn}. For the engagement traces of GameVibe, we adopt the Gaussian distribution following the prior works \cite{zhang2017predicting, atcheson2019using}. The distribution parameter values are set using the maximum likelihood estimation as described in Section \ref{sec:method}. 

\subsection{Emotion Model Settings} 

The LSTM module is a two-layer, unidirectional network that processes sequential input representations, followed by a linear transformation with a $tanh$ activation function, applied at each time step, to generate the final outputs. 
For RECOLA, the training utterances are split into 100 frames in each batch to improve training efficiency and tested with the entire utterance to match the practical scenarios. For GameVibe, the videos are split into 19 time windows in each batch to match the corpus' video length. We use the Adam optimizer to train our LSTM models with an initial learning rate of 1e-3 and a weight decay of 1e-4.
For RECOLA, the maximum number of iterations is set to 100, with early stopping based on the best loss performance on the test set. For GameVibe, the maximum number of iterations is set to 2,500 using the same early stopping method. Training is guided by the Concordance Correlation Coefficient (CCC) loss given as $L_\rho = 1 - \rho$, where $\rho$ is the CCC value \cite{lawrence1989concordance}.

\subsection{Evaluation Measures}

Emotion models are evaluated using both the CCC \cite{lawrence1989concordance} and the Signed Differential Agreement (SDA) \cite{booth2020fifty} measures. CCC is a conventional metric commonly used in continuous emotion recognition tasks. We selected SDA as a complementary metric to CCC because it is a pairwise measure of reliability between two signals, emphasizing the alignment of their directional trends rather than their magnitude. This property makes SDA particularly well-suited for assessing ordinal labels.

\section{Results}

In this section, we evaluate the impact of different emotion representations used in a traditional CER system by testing the proposed representations on the RECOLA and GameVibe datasets.

\subsection{RECOLA} \label{sec:recola}

The results presented in Table \ref{table:bounded} show that the interval representation $I$ yields the highest performance in terms of CCC values for both $\mu$ and $\sigma$ predictions. However, the proposed $O^\text{G}$ and $O^\text{I}$ representations perform better when evaluated via the SDA metric. For instance, the $\mu$ predictions of arousal show that $O^\text{G}$ outperforms $I$ with a relative increase of $17.67\%$, and $O^\text{I}$ outperforms it by $12.09\%$. A similar pattern is observed for the $\mu$ prediction of valence, where $O^\text{I}$ (0.221) and $O^\text{G}$ (0.219) achieves higher performances than $I$ (0.130). This suggests that while $I$ is more effective for capturing absolute emotion states, ordinal representations better reflect relative changes in emotion perception, as SDA specifically measures agreement in directional changes rather than magnitude. We do not observe consistent results for the $\sigma$ predictions, likely due to the common challenges associated with predicting $\sigma$, as noted in earlier research \cite{han2017hard, wu2024dual}.

Interestingly, we also note that $O^\text{G}$ outperforms $O^\text{I}$ for arousal prediction (both $\mu$ and $\sigma$) in terms of both CCC and SDA measures. Specifically, $O^\text{G}$ outperforms $O^\text{I}$ by a relative increase of $4.98\%$ for $\mu$ prediction while, for $\sigma$ prediction, $O^\text{G}$ (0.286) demonstrates higher predictive capacity than $O^\text{I}$ (0.034). These findings indicate that capturing the temporal evolution of emotion ambiguity at the group level provides a more robust and reliable representation than modeling individual annotators' temporal dynamics separately. Directly tracking how the group distribution changes likely smooth out individual inconsistencies, leading to better alignment with the overall emotional trajectory. Although we do not observe consistent results on valence---as predicting valence from speech features is generally more challenging \cite{wu2021multimodal, tzirakis2019real, bachorowski1999vocal}---the predictions for $\mu$ on valence still yield high CCC values and remained comparable in terms of SDA values. 

\begin{table}[!tb]
\caption{RECOLA: CCC and SDA performance across emotion representations.}\vspace{-0.5em}
    \centering   
    \begin{tabular}{ccccc}
    \multicolumn{5}{c}{\textbf{Arousal}} \\
 \hline
       \textbf{Representation} & \textbf{CCC $\mu$} & \textbf{CCC $\sigma$} & \textbf{SDA $\mu$} & \textbf{SDA $\sigma$} \\
       \hline
        {$I$}       & \textbf{0.759}  & \textbf{0.466} & 0.430 & 0.286  \\
        {$O^\text{I}$}  & 0.604  & 0.127 & 0.482 & 0.034 \\
        {$O^\text{G}$} & {0.726}  & {0.354} & \textbf{0.506} & \textbf{0.286} \\
        
        \hline
        \multicolumn{5}{c}{\textbf{Valence}} \\
        \hline
        \textbf{Representation} & \textbf{CCC $\mu$} & \textbf{CCC $\sigma$} & \textbf{SDA $\mu$} & \textbf{SDA $\sigma$} \\
       \hline
        {$I$}       & \textbf{0.471}  & \textbf{0.225} & 0.130 & -0.001  \\
        {$O^\text{I}$}  & 0.197  & 0.178 & \textbf{0.221} & 0.059 \\
        {$O^\text{G}$} & {0.397}  & 0.056 & {0.219} & \textbf{0.077} \\
        \hline
    \end{tabular}
    \label{table:bounded}
\end{table}

\subsection{GameVibe}\label{sec:gamevibe:raw}

\begin{table}[tb!]
\caption{GameVibe (raw): CCC and SDA performance across emotion representations.}\vspace{-0.5em}
    \centering
    \resizebox{\columnwidth}{!}{
    \begin{tabular}{ccccc}
    \hline
       \textbf{Representation} & \textbf{CCC $\mu$} & \textbf{CCC $\sigma$} & \textbf{SDA $\mu$} & \textbf{SDA $\sigma$} \\
       \hline
        {$I$}       & $0.055\pm0.034$  & $0.033\pm0.006$ & $0.223\pm0.090$& $\bm{0.586\pm0.058}$  \\
        {$O^\text{I}$}  & $0.135\pm0.067$  & $0.094\pm0.021$ & $0.223\pm0.057$& $0.534\pm0.056$ \\
        {$O^\text{G}$} & \textbf{$\bm{0.250\pm0.065}$}  & $\bm{0.173\pm0.068}$ & \textbf{$\bm{0.297\pm0.061}$} & $0.112\pm0.031$ \\
        \hline

    \end{tabular}
    }

    \label{tab:gamevibe:raw}
\end{table}

Table \ref{tab:gamevibe:raw} shows the results of our experiments on the GameVibe unbounded ordinal dataset. Because GameVibe data is unbounded and ordinal, engagement traces can virtually range infinitely from any negative to any positive value. Looking at $\mu$ predictions, we can observe that $O^\text{G}$ outperforms the other types of emotion representations both in terms of CCC ($0.250$) and SDA ($0.297$). These findings confirm that the rate of change across multiple annotators is a more robust representation of emotion, offering greater reliability. Aggregating trends across multiple annotators reduces noise and enhances predictive accuracy---at least when it comes to the central tendency of the signal ($\mu$). These results also benchmark the performances of performing regression tasks on this dataset - which have never done before.

\begin{figure}[!tb]
\centering
\includegraphics[width=1.0\linewidth]{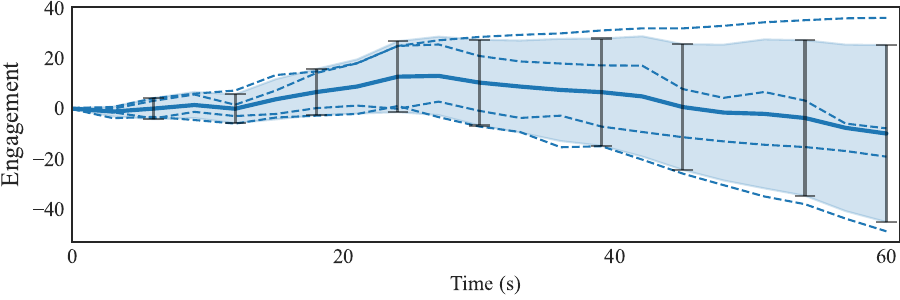}\vspace{-0.5mm}
\caption{An example of signal dissociation over time between annotators in the GameVibe dataset. The dashed lines show the trace of each individual annotator; the sold line shows the central tendency of the annotation ($\mu$); the shaded area and error bars show the standard deviation over time ($\sigma$).}
\label{fig:dissociation}
\vspace{-1.5mm}
\end{figure}

The picture is not immediately clear, however, when we look at predictions of ambiguity ($\sigma$). While $O^\text{G}$ still offers the best representation in terms of CCC ($0.173$), the obtained SDA values reveal the opposite finding. Particularly, $O^\text{G}$ yields the lowest SDA value ($0.112$), while $I$ 
demonstrates a staggering SDA value of $0.586$, which indicates considerable agreement with the deviation existent in the ground truth. This occurs because, there is a tendency of \emph{unbounded} human annotation traces to progressively diverge over time (see an example on Fig.~\ref{fig:dissociation}). As a result, signal variations increase significantly, making it easier to predict the values of $\sigma$ but leading to a loss of sensitivity in tracking changes in $\sigma$.

\section{Discussion \& Conclusions}

This paper introduced a novel paradigm for emotion representation we name \emph{Ambiguity-aware Ordinal Emotional Representation}. This paradigm is demonstrated to reliably capture both the ambiguity and the temporal dynamics of emotion traces. We tested the proposed representations on two well-established datasets---RECOLA and GameVibe---using both bounded interval (arousal and valence) and unbounded ordinal (engagement ) continuous labels. Our results show that even though using the absolute values of time-continuous labels are beneficial when modeling bounded annotations---an ordinal representation is beneficial at modeling unbounded traces and estimating the temporal dynamics of the data both in terms of central tendency and ambiguity. Furthermore, our findings indicate that our proposed \emph{Group} representation outperforms \emph{Individual} representation, underscoring the importance of relying on global trends in modeling emotional ambiguity. By integrating these representations into conventional affect modeling systems, we demonstrate that detecting emotions through their relative temporal changes---rather than through their static magnitudes---enhances the model's ability to better align with human perception.

\setlength{\parskip}{0pt}
While initial findings suggest that capturing ambiguity via ordinal representations is promising, there are a number of limitations related to our study. Most notably, we approach the modeling of the central tendency and the ambiguity of the annotation traces as a regression downstream task. Although continuous affect labels and emotional intensity are often modeled in such a way \cite{toisoul2021estimation, kollias2023abaw}, this approach does not adhere to natural cognitive processes---which are inherently ordinal \cite{yannakakis2018ordinal}. Future work should focus on approaching the problem through alternative learning paradigms such as classification \cite{melhart2020moment} and preference learning \cite{melhart2021towards}. Our results on GameVibe highlight the sensitivity of the models to global trends in the data. As unbounded labels tend to dissociate over time, future work should investigate the effect of data normalization on our results. Beyond the different normalization strategies we could employ, another way to increase the ecological validity of our results is to reproduce them across new datasets and modalities. While some limited experiments on building general affect models have already been reported in the literature \cite{camilleri2017towards, melhart2021towards}, our results on \textit{Ambiguity-aware Ordinal Group Representation} open new avenues for future research in this area. Further work in the future should also integrate and test ambiguity-aware ordinal emotion representation with alternative models beyond LSMTs such as transformer-based architectures.

Additionally, while ordinal representations are effective for modeling relative changes in emotion perception, they may be less suited for applications that require capturing absolute emotional intensity (e.g., identifying spikes of "high arousal" states).
Furthermore, although we evaluated interval and ordinal representations separately, combining them as complementary inputs may offer a more holistic view—--particularly by capturing both absolute levels and relative changes in emotions. Future work could explore hybrid architectures that jointly leverage the strengths of both representations to enhance continuous affect modeling, especially under ambiguous or subjective annotation conditions.

In summary, our study highlights the significance of temporal dynamics and inter-rater ambiguity in emotion modeling, and introduces a novel approach to emotion representation. Through a series of experiments, we showcased different methods of affect-aware emotion representations and demonstrated the strengths of ordinal approaches when it comes to estimating central trends and the changing ambiguity of annotations---paving the way for more robust and human-aligned affective computing systems.

\section{Ethical Impact Statement}
This work introduces a novel ambiguity-aware ordinal representation to model both the ambiguous and ordinal nature of emotions, emphasizing the role of temporal changes in affective computing. Unlike conventional methods that treat emotional states as static or independent variables, our approach leverages ordinal structures to better capture the progression and relative changes in emotion over time. By accounting for variations in emotion annotation and tracing how emotions evolve, our framework provides a more structured and interpretable representation of human emotions.

The ethical implications of modeling emotions—and the inherent challenges in capturing their ambiguous and ordinal nature—are significant, especially in applications related to mental health, human-computer interaction, and affective AI. Misrepresenting or oversimplifying emotional states can lead to misleading conclusions and unintended consequences. Our approach mitigates such risks by emphasizing relational and progression-based representations, which better reflect the way emotions are perceived and experienced.

Despite these advantages, there are potential risks associated with deploying ordinal emotion models in real-world applications. We acknowledge that even with the proposed representation, our system is trained on given datasets collected from two specific group of annotators. This highlights the importance of ensuring that such systems are trained and evaluated across diverse populations to prevent reinforcing existing disparities or introducing unintended limitations. Therefore, it is essential to for improving the generalizability and fairness of affective computing models.

Another critical ethical consideration involves privacy concerns. Affective computing systems, including the methods proposed in this work, require emotion data collected from human participants, which may pose privacy and surveillance risks. To safeguard individuals’ rights, it is essential to implement stringent privacy protections, including secure data collection, storage, and processing practices. Clear regulatory frameworks should be established to ensure ethical data usage, and users must be fully informed about the nature of the data being collected while providing explicit consent before participation.

\IEEEtriggeratref{38}
\bibliographystyle{IEEEtran}

\bibliography{references}

\begin{thebibliography}{10}
\providecommand{\url}[1]{#1}
\csname url@samestyle\endcsname
\providecommand{\newblock}{\relax}
\providecommand{\bibinfo}[2]{#2}
\providecommand{\BIBentrySTDinterwordspacing}{\spaceskip=0pt\relax}
\providecommand{\BIBentryALTinterwordstretchfactor}{4}
\providecommand{\BIBentryALTinterwordspacing}{\spaceskip=\fontdimen2\font plus
\BIBentryALTinterwordstretchfactor\fontdimen3\font minus \fontdimen4\font\relax}
\providecommand{\BIBforeignlanguage}[2]{{%
\expandafter\ifx\csname l@#1\endcsname\relax
\typeout{** WARNING: IEEEtran.bst: No hyphenation pattern has been}%
\typeout{** loaded for the language `#1'. Using the pattern for}%
\typeout{** the default language instead.}%
\else
\language=\csname l@#1\endcsname
\fi
#2}}
\providecommand{\BIBdecl}{\relax}
\BIBdecl

\bibitem{wang2022systematic}
Y.~Wang, W.~Song, W.~Tao, A.~Liotta, D.~Yang, X.~Li, S.~Gao, Y.~Sun, W.~Ge, W.~Zhang \emph{et~al.}, ``A systematic review on affective computing: Emotion models, databases, and recent advances,'' \emph{Information Fusion}, 2022.

\bibitem{ekman1999basic}
P.~Ekman \emph{et~al.}, ``Basic emotions,'' \emph{Handbook of cognition and emotion}, vol.~98, no. 45-60, p.~16, 1999.

\bibitem{russell1980circumplex}
J.~A. Russell, ``A circumplex model of affect.'' \emph{Journal of personality and social psychology}, vol.~39, no.~6, 1980.

\bibitem{barthet2024gamevibe}
M.~Barthet, M.~Kaselimi, K.~Pinitas, K.~Makantasis, A.~Liapis, and G.~N. Yannakakis, ``{GameVibe}: a multimodal affective game corpus,'' \emph{Scientific Data}, vol.~11, no.~1, p. 1306, 2024.

\bibitem{ringeval2013introducing}
F.~Ringeval, A.~Sonderegger, J.~Sauer, and D.~Lalanne, ``Introducing the recola multimodal corpus of remote collaborative and affective interactions,'' in \emph{2013 10th IEEE international conference and workshops on automatic face and gesture recognition (FG)}.\hskip 1em plus 0.5em minus 0.4em\relax IEEE, 2013, pp. 1--8.

\bibitem{sethu2019ambiguous}
V.~Sethu, E.~M. Provost, J.~Epps, C.~Busso, N.~Cummins, and S.~Narayanan, ``The ambiguous world of emotion representation,'' \emph{arXiv preprint arXiv:1909.00360}, 2019.

\bibitem{gunes2010automatic}
H.~Gunes and M.~Pantic, ``Automatic, dimensional and continuous emotion recognition,'' \emph{International Journal of Synthetic Emotions (IJSE)}, vol.~1, no.~1, pp. 68--99, 2010.

\bibitem{gunes2013categorical}
H.~Gunes and B.~Schuller, ``Categorical and dimensional affect analysis in continuous input: Current trends and future directions,'' \emph{Image and Vision Computing}, vol.~31, no.~2, pp. 120--136, 2013.

\bibitem{ringeval2015avec}
F.~Ringeval, B.~Schuller, M.~Valstar, R.~Cowie, and M.~Pantic, ``{AVEC} 2015: The 5th international audio/visual emotion challenge and workshop,'' in \emph{Proceedings of the 23rd ACM international conference on Multimedia}, 2015, pp. 1335--1336.

\bibitem{tzirakis2018end}
P.~Tzirakis, J.~Zhang, and B.~W. Schuller, ``End-to-end speech emotion recognition using deep neural networks,'' in \emph{2018 IEEE international conference on acoustics, speech and signal processing (ICASSP)}.\hskip 1em plus 0.5em minus 0.4em\relax IEEE, 2018, pp. 5089--5093.

\bibitem{han2017hard}
J.~Han, Z.~Zhang, M.~Schmitt, M.~Pantic, and B.~Schuller, ``From hard to soft: Towards more human-like emotion recognition by modelling the perception uncertainty,'' in \emph{Proceedings of the 25th ACM international conference on Multimedia}, 2017, pp. 890--897.

\bibitem{dang2018dynamic}
T.~Dang, V.~Sethu, and E.~Ambikairajah, ``Dynamic multi-rater gaussian mixture regression incorporating temporal dependencies of emotion uncertainty using kalman filters,'' in \emph{2018 IEEE International Conference on Acoustics, Speech and Signal Processing (ICASSP)}.\hskip 1em plus 0.5em minus 0.4em\relax IEEE, 2018, pp. 4929--4933.

\bibitem{bose2021parametric}
D.~Bose, V.~Sethu, and E.~Ambikairajah, ``Parametric distributions to model numerical emotion labels,'' \emph{Proc. Interspeech 2021}, pp. 4498--4502, 2021.

\bibitem{wu2022novel}
J.~Wu, T.~Dang, V.~Sethu, and E.~Ambikairajah, ``A novel sequential {Monte Carlo} framework for predicting ambiguous emotion states,'' in \emph{ICASSP 2022-2022 IEEE International Conference on Acoustics, Speech and Signal Processing (ICASSP)}.\hskip 1em plus 0.5em minus 0.4em\relax IEEE, 2022, pp. 8567--8571.

\bibitem{bose2024continuous}
D.~Bose, V.~Sethu, and E.~Ambikairajah, ``Continuous emotion ambiguity prediction: Modeling with beta distributions,'' \emph{IEEE Transactions on Affective Computing}, 2024.

\bibitem{wu2024can}
Y.-T. Wu, J.~Wu, V.~Sethu, and C.-C. Lee, ``Can modelling inter-rater ambiguity lead to noise-robust continuous emotion predictions?'' in \emph{Proc. Interspeech 2024}, 2024, pp. 3714--3718.

\bibitem{wu2024dual}
J.~Wu, T.~Dang, V.~Sethu, and E.~Ambikairajah, ``Dual-constrained dynamical neural odes for ambiguity-aware continuous emotion prediction,'' \emph{arXiv preprint arXiv:2407.21344}, 2024.

\bibitem{yannakakis2018ordinal}
G.~N. Yannakakis, R.~Cowie, and C.~Busso, ``The ordinal nature of emotions: An emerging approach,'' \emph{IEEE Transactions on Affective Computing (Early Access)}, 2018.

\bibitem{yannakakis2017ordinal}
------, ``The ordinal nature of emotions,'' in \emph{Proceedings of the Intl. Conference on Affective Computing and Intelligent Interaction}.\hskip 1em plus 0.5em minus 0.4em\relax IEEE, 2017, pp. 248--255.

\bibitem{junge2013indirect}
M.~Junge and R.~Reisenzein, ``Indirect scaling methods for testing quantitative emotion theories,'' \emph{Cognition \& Emotion}, vol.~27, no.~7, pp. 1247--1275, 2013.

\bibitem{laming1984relativity}
D.~Laming, ``The relativity of ‘absolute’judgements,'' \emph{British Journal of Mathematical and Statistical Psychology}, vol.~37, no.~2, pp. 152--183, 1984.

\bibitem{wu2023interval}
J.~Wu, T.~Dang, V.~Sethu, and E.~Ambikairajah, ``From interval to ordinal: A {HMM} based approach for emotion label conversion,'' in \emph{Proc. Interspeech 2023}, 2023, pp. 1843--1847.

\bibitem{phillips2018impact}
C.~Phillips, D.~Johnson, M.~Klarkowski, M.~J. White, and L.~Hides, ``The impact of rewards and trait reward responsiveness on player motivation,'' in \emph{Proceedings of the 2018 Symposium on Computer-Human Interaction in Play}.\hskip 1em plus 0.5em minus 0.4em\relax ACM, 2018, pp. 393--404.

\bibitem{damasio1994descartes}
A.~R. Damasio, \emph{Descartes’ error: Emotion, rationality and the human brain}.\hskip 1em plus 0.5em minus 0.4em\relax New York: Putnam, 1994.

\bibitem{seymour2008anchors}
B.~Seymour and S.~M. McClure, ``Anchors, scales and the relative coding of value in the brain,'' \emph{Current opinion in neurobiology}, vol.~18, no.~2, pp. 173--178, 2008.

\bibitem{johnson2005relation}
T.~Johnson, P.~Kulesa, Y.~I. Cho, and S.~Shavitt, ``The relation between culture and response styles: Evidence from 19 countries,'' \emph{Journal of Cross-cultural psychology}, vol.~36, no.~2, pp. 264--277, 2005.

\bibitem{tversky1981framing}
A.~Tversky and D.~Kahneman, ``The framing of decisions and the psychology of choice,'' \emph{Science}, vol. 211, no. 4481, pp. 453--458, 1981.

\bibitem{melhart2020study}
D.~Melhart, K.~Sfikas, G.~Giannakakis, and G.~Y.~A. Liapis, ``A study on affect model validity: Nominal vs ordinal labels,'' in \emph{Workshop on Artificial Intelligence in Affective Computing}.\hskip 1em plus 0.5em minus 0.4em\relax Proceedings of Machine Learning Research, 2020, pp. 27--34.

\bibitem{zhang2018dynamic}
Z.~Zhang, J.~Han, E.~Coutinho, and B.~Schuller, ``Dynamic difficulty awareness training for continuous emotion prediction,'' \emph{IEEE Transactions on Multimedia}, vol.~21, no.~5, pp. 1289--1301, 2018.

\bibitem{huang2015investigation}
Z.~Huang, T.~Dang, N.~Cummins, B.~Stasak, P.~Le, V.~Sethu, and J.~Epps, ``An investigation of annotation delay compensation and output-associative fusion for multimodal continuous emotion prediction,'' in \emph{Proceedings of the 5th International Workshop on Audio/Visual Emotion Challenge}, 2015, pp. 41--48.

\bibitem{lopes2017ranktrace}
P.~Lopes, G.~N. Yannakakis, and A.~Liapis, ``Ranktrace: Relative and unbounded affect annotation,'' in \emph{Proceedings of the Intl. Conference on Affective Computing and Intelligent Interaction}.\hskip 1em plus 0.5em minus 0.4em\relax IEEE, 2017, pp. 158--163.

\bibitem{melhart2019pagan}
D.~Melhart, A.~Liapis, and G.~N. Yannakakis, ``{PAGAN}: Video affect annotation made easy,'' in \emph{Proceedings of the Conference on Affective Computing and Intelligent Interaction (ACII)}, 2019.

\bibitem{pinitas2024varying}
K.~Pinitas, N.~Rasajski, M.~Barthet, M.~Kaselimi, K.~Makantasis, A.~Liapis, and G.~N. Yannakakis, ``Varying the context to advance affect modelling: A study on game engagement prediction,'' in \emph{Proceedings of the Conference on Affective Computing and Intelligent Interaction (ACII)}, 2024.

\bibitem{pinitas2024across}
K.~Pinitas, K.~Makantasis, and G.~N. Yannakakis, ``Across-game engagement modelling via few-shot learning,'' \emph{arXiv preprint arXiv:2409.13002}, 2024.

\bibitem{schmitt2016border}
M.~Schmitt, F.~Ringeval, and B.~W. Schuller, ``At the border of acoustics and linguistics: Bag-of-audio-words for the recognition of emotions in speech.'' in \emph{Interspeech}, 2016, pp. 495--499.

\bibitem{schmitt2017openxbow}
M.~Schmitt and B.~Schuller, ``openxbow--introducing the passau open-source crossmodal bag-of-words toolkit,'' \emph{Journal of Machine Learning Research}, vol.~18, no.~96, pp. 1--5, 2017.

\bibitem{wang2023videomae}
L.~Wang, B.~Huang, Z.~Zhao, Z.~Tong, Y.~He, Y.~Wang, Y.~Wang, and Y.~Qiao, ``Videomae v2: Scaling video masked autoencoders with dual masking,'' in \emph{Proceedings of the IEEE/CVF conference on computer vision and pattern recognition}, 2023, pp. 14\,549--14\,560.

\bibitem{zhang2017predicting}
B.~Zhang, G.~Essl, and E.~Mower~Provost, ``Predicting the distribution of emotion perception: capturing inter-rater variability,'' in \emph{Proceedings of the 19th ACM International Conference on Multimodal Interaction}, 2017, pp. 51--59.

\bibitem{atcheson2019using}
M.~Atcheson, V.~Sethu, and J.~Epps, ``Using gaussian processes with lstm neural networks to predict continuous-time, dimensional emotion in ambiguous speech,'' in \emph{2019 8th International Conference on Affective Computing and Intelligent Interaction (ACII)}.\hskip 1em plus 0.5em minus 0.4em\relax IEEE, 2019, pp. 718--724.

\bibitem{lawrence1989concordance}
I.~Lawrence and K.~Lin, ``A concordance correlation coefficient to evaluate reproducibility,'' \emph{Biometrics}, pp. 255--268, 1989.

\bibitem{booth2020fifty}
B.~M. Booth and S.~S. Narayanan, ``Fifty shades of green: Towards a robust measure of inter-annotator agreement for continuous signals,'' in \emph{Proceedings of the 2020 international conference on multimodal interaction}, 2020, pp. 204--212.

\bibitem{wu2021multimodal}
J.~Wu, T.~Dang, V.~Sethu, and E.~Ambikairajah, ``Multimodal affect models: An investigation of relative salience of audio and visual cues for emotion prediction,'' \emph{Frontiers in Computer Science}, vol.~3, p. 767767, 2021.

\bibitem{tzirakis2019real}
P.~Tzirakis, S.~Zafeiriou, and B.~Schuller, ``Real-world automatic continuous affect recognition from audiovisual signals,'' in \emph{Multimodal behavior analysis in the wild}.\hskip 1em plus 0.5em minus 0.4em\relax Elsevier, 2019, pp. 387--406.

\bibitem{bachorowski1999vocal}
J.-A. Bachorowski, ``Vocal expression and perception of emotion,'' \emph{Current directions in psychological science}, vol.~8, no.~2, pp. 53--57, 1999.

\bibitem{toisoul2021estimation}
A.~Toisoul, J.~Kossaifi, A.~Bulat, G.~Tzimiropoulos, and M.~Pantic, ``Estimation of continuous valence and arousal levels from faces in naturalistic conditions,'' \emph{Nature Machine Intelligence}, vol.~3, no.~1, pp. 42--50, 2021.

\bibitem{kollias2023abaw}
D.~Kollias, P.~Tzirakis, A.~Baird, A.~Cowen, and S.~Zafeiriou, ``Abaw: Valence-arousal estimation, expression recognition, action unit detection \& emotional reaction intensity estimation challenges,'' in \emph{Proceedings of the IEEE/CVF Conference on Computer Vision and Pattern Recognition}, 2023, pp. 5889--5898.

\bibitem{melhart2020moment}
D.~Melhart, D.~Gravina, and G.~N. Yannakakis, ``Moment-to-moment engagement prediction through the eyes of the observer: Pubg streaming on twitch,'' in \emph{Proceedings of the 15th International Conference on the Foundations of Digital Games}, 2020, pp. 1--10.

\bibitem{melhart2021towards}
D.~Melhart, A.~Liapis, and G.~N. Yannakakis, ``Towards general models of player experience: A study within genres,'' in \emph{2021 IEEE Conference on Games (CoG)}.\hskip 1em plus 0.5em minus 0.4em\relax IEEE, 2021, pp. 01--08.

\bibitem{camilleri2017towards}
E.~Camilleri, G.~N. Yannakakis, and A.~Liapis, ``Towards general models of player affect,'' in \emph{Proceedings of the Intl. Conference on Affective Computing and Intelligent Interaction}.\hskip 1em plus 0.5em minus 0.4em\relax IEEE, 2017, pp. 333--339.

\end{thebibliography}

\end{document}